\documentclass[a4paper,fleqn]{cas-dc}

\usepackage[authoryear,longnamesfirst]{natbib}

\def\tsc#1{\csdef{#1}{\textsc{\lowercase{#1}}\xspace}}
\tsc{WGM}
\tsc{QE}

\usepackage{graphicx}
\usepackage{amsmath}
\usepackage{amssymb}
\usepackage{booktabs,siunitx}
\usepackage{multirow}
\usepackage[utf8]{inputenc}
\usepackage{pgfplots}
\DeclareUnicodeCharacter{2212}{−}
\usepgfplotslibrary{groupplots,dateplot}
\usetikzlibrary{patterns,shapes.arrows}
\pgfplotsset{compat=newest}
\usepackage{caption}
\usepackage{subcaption}
\usepackage{tikz}

\begin{document}
\let\WriteBookmarks\relax
\def\floatpagepagefraction{1}
\def\textpagefraction{.001}

\shorttitle{KGNT-ens: Few-Shot Image Classification with Knowledge Graph Ensembles}    

\shortauthors{D. Filipiak, A. Fensel, A. Filipowska}  

\title [mode = title]{KGTN-ens: Few-Shot Image Classification with Knowledge Graph Ensembles}  



\author[uibk,uw]{Dominik Filipiak}[orcid=0000-0002-4927-9992]
\credit{conceptualization, data curation, formal analysis, investigation, methodology, software, validation, writing -- original draft (85\% of the work in total)}
\cormark[1]
\cortext[mycorrespondingauthor]{Corresponding author}
\ead{dfilipiak@mimuw.edu.pl}
\author[wur,uibk]{Anna Fensel}[orcid=0000-0002-1391-7104]
\credit{conceptualization, funding acquisition, project administration, writing -- review \& editing (10\% of the work in total)}
\ead{anna.fensel@wur.nl}
\author[uep]{Agata Filipowska}[orcid=0000-0002-8425-1872]
\credit{writing -- review \& editing, resources (5\% of the work in total)}
\ead{agata.filipowska@ue.poznan.pl}
\affiliation[uibk]{organization={University of Innsbruck},
            addressline={Innrain 52}, 
            city={Innsbruck},
            postcode={6020}, 
            country={Austria}}
\affiliation[uw]{organization={University of Warsaw},
            addressline={Krakowskie Przedmieście 26/28}, 
            city={Warsaw},
            postcode={00-927}, 
            country={Poland}}
\affiliation[wur]{organization={Wageningen University \& Research},
            addressline={Droevendaalsesteeg 2}, 
            city={Wageningen},
            postcode={6708 PB}, 
            country={The Netherlands}}
\affiliation[uep]{organization={Pozna\'n University of Economics and Business},
            addressline={Al. Niepodległości 10}, 
            city={Poznań},
            postcode={61-875}, 
            country={Poland}}


\begin{abstract}
  We propose KGTN-ens, a framework extending the recent Knowledge Graph Transfer Network (KGTN) to be able to incorporate multiple knowledge graph embeddings at a small cost.
  We evaluate it with different combinations of embeddings in a few-shot image classification task.
  We also construct a new knowledge source -- Wikidata embeddings --  and evaluate it with KGTN and KGTN-ens.
  Our approach outperforms KGTN in terms of the top-5 accuracy on the ImageNet-FS dataset for the majority of tested settings.
  The code is available on GitHub: \texttt{The code will be released after the publication}.
\end{abstract}



\begin{keywords}
  Few-shot Image Classification
  \sep 
  Knowledge Graph Enabled AI
  \sep
  Ensemble Learning
\end{keywords}

\maketitle

\section{Introduction}
\label{sec:introduction}

Deep learning has made a substantial impact on a number of industrial and research areas.
This includes computer vision, as the rapid development of representation learning started with the seminal work of \cite{krizhevsky2017imagenet} for the image classification task.
However, numerous state-of-the-art models often require large amounts of data to train, which can be costly to gather and label -- especially for vision-related tasks.
Therefore, an intense research effort can be observed in the area of data-efficient machine learning methods.
Few-shot learning (often abbreviated as FSL) is a machine learning task, where the machine learning model is (partially) trained on a small amount of data -- part of the labelled data is available in standard amounts, whereas the other part consists of only a few (typically less than 10) samples per class.
Few-shot learning can also suffer from selection bias since the decision boundaries need to be adjusted to a new few samples, which can contain irrelevant and misleading artefacts (such as a background colour).
Hence the learning process is substantially more challenging.

One way to tackle the few-shot learning task is to use some prior knowledge of the labelled data.
Knowledge Graph Transfer Network (KGTN), the recent work of \cite{chen2020knowledge}, solves this problem by learning the prototypes from the external sources of knowledge and comparing them against extracted features from an input image.
A similarity function scores the output of these two and yields the class probability distribution. 
These external sources of knowledge are represented as class correlation matrices.
A vital element of this architecture is the knowledge graph transfer module (KGTM), which tries to learn class prototypes from knowledge graph embeddings using gated graph neural networks (GGTN) \citep{li2016gated}.

In the KGTN approach, one has to select a single knowledge source.
Inspired by ensemble learning approaches, this observation leads us to the following questions: is it possible to learn prototypes from multiple knowledge graph embeddings?
If so, will it result in higher performance metrics values, such as accuracy for classification problems?
Therefore, we propose KGTN-ens, an extension of KGTN, that use multiple embeddings instead of a single one. 
Each of them generates different prototypes, which are later combined and compared against the output of the feature extractor. 
We test two ensemble learning techniques in this paper.
We also evaluated different combinations of three knowledge graphs, one of which (based on Wikidata) is introduced by us and has not been used in the original paper.
Our solution is knowledge graph agnostic, provided that the knowledge graph is embedded and linked to the classes used in the image classification.

The contribution of this paper is two-fold: (1)~we propose KGTN-ens, a new method based on KGTN, and evaluate it with different combinations of embeddings, (2)~we construct a new knowledge source -- Wikidata embeddings --  and evaluate it with KGTN and KGTN-ens.
Our approach outperforms KGTN in terms of the top-5 accuracy on the ImageNet-FS dataset for the majority of tested settings.

The remainder of this paper is organised as follows.
A~comprehensive literature survey on related work is presented in Section \ref{sec:related_work}.
Section \ref{sec:method} provides a description of the KGTN-ens architecture.
Section \ref{sec:results} describes the results of the evaluation of Wikidata embeddings with KGTN and KGTN-ens with different combinations of embeddings, along with the detailed analysis and ablation studies.
Section \ref{sec:conclusion} concludes the paper.

\section{Related work}
\label{sec:related_work}

This section provides a comprehensive overview of the related work.
We start with a brief review of the techniques used for graph neural networks, which are at the core of the KGTN-ens architecture.
Then, we provide a short survey on recent advancements in few-shot learning, which is the main machine learning task solved by the architecture presented in this paper.

\textbf{Graph neural networks.}
In general, Graph neural networks (GNNs) represent a type of neural network, which processes the specified attributes of the graphs.
Task tackled by GNN can be either node-level (such as prediction of a property for each node), edge-level (prediction of a property for each edge), or graph-level (prediction of a property for a whole graph) \citep{sanchez-lengeling2021a}.
Following \cite[]{keriven2019universal}, a crucial feature of GNNs is being either invariant or equivariant to permutations.
That is, for a graph $\mathcal{G}$, network $f$ and a permutation $\Pi$ we have $f(\Pi \star \mathcal{G})=f(\mathcal{G})$ and $f(\Pi \star \mathcal{G})=\Pi \star f(\mathcal{G})$ for invariance and equivariance respectively.
The general-purpose models from the state-of-the-art family of transformer architectures \cite{vaswani2017attention} can be viewed as a special instance of a graph neural network.
Graph neural networks have a wide area of applications, with notable examples in biology (e.g. protein interface prediction) or social networks (e.g. community detection or link prediction).
The less obvious application of GNNs is in the field of image classification, where they are used to learn the prototypes from the knowledge graph embeddings in a few-shot learning setting.

GNNs fall into a broader category of geometric deep learning, which is devoted to the application of deep neural networks on structured non-Euclidean domains, such as graphs, manifolds, meshes, or grids \citep{bronstein2017geometric}.
\cite{gilmer2017neural} proposed \emph{message passing}, which is one of the most important concepts in GNNs.
In this approach, nodes and/or edges can rely on their neighbours in order to create meaningful embeddings iteratively.
\cite{wu2020comprehensive} classify GNNs into four broad categories: recurrent GNNs (RecGNN), convolutional GNNs (ConvGNNs), graph autoencoders (GAEs), and spatial-temporal GNNs (STGNNs).
In this article, Gated Graph Neural Network (GGNN) \citep{li2016gated} are of special interest.
They belong to the category of RecGNNs.
For a fair comparison with KGTN (our baseline), we used GGNN in our experiments.
Following \cite{li2016gated}, the intuitive difference between GNN and GGNN relies on the explicit graph structure of GNNs, which results in more generalisation capabilities at the expense of a less general model of the latter.

\textbf{Few-shot learning.}
While being very effective for numerous vision tasks, one of the main problems with convolutional neural networks (or machine learning in general) is the amount of data they need to provide meaningful predictions.
More recent architectures, such as self-attention models require even more data to train.
On contrary, humans typically require only a few samples to acquire knowledge of seen objects.
One way to tackle this issue is few-shot learning, which is aimed at learning from scarce data.
The complexity of the problem often stems from the required sudden shift of decision boundaries, which is hard to achieve using only a few samples.
A special case of few-shot learning is one-shot learning, which is learning from one labelled sample per class.

Following \cite{song2022comprehensive}, few-shot learning methods can be divided into data augmentation, transfer learning, meta-learning, and multimodal learning.
Data augmentation techniques aim to artificially extend the amount of available data by either transforming input data \citep{chen2019image} or resulting features \citep{chen2019multi}.
Transfer learning focuses on resuing features from networks trained on different datasets with the required amount of data by techniques such as pre-training and fine-tuning or domain adaptation.
Meta-learning includes techniques devoted to learning from data and tasks in order to reuse this knowledge for future downstream tasks.
\cite{finn2017model} proposed a model agnostic meta-learning algorithm MAML.
Specialised approaches to meta-learning include neural architecture search \citep{elsken2019neural} or metric learning \citep{ge2018deep,chicco2021siamese}.
Finally, multimodal learning focuses on the incorporation of external knowledge from heterogenous domains, such as text, speech or knowledge graphs \citep{wang2020large}.

The concept of prototypes was introduced in the work of \cite{snell2017prototypical}, where they proposed prototypical networks focused on learning metric space between class instances and their prototypes.
\cite{hariharan2017low} used representation regularisation and introduced the concept of \emph{hallucinations} in order to enlarge the number of available representations during the training.  
\cite{wang2018low} employed meta-learning techniques and combined them with the aforementioned \emph{hallucinations} to improve few-shot classification metrics.
A growing number of scholars incorporate structured knowledge into their computer vision research \cite[]{monka2022survey}.
For instance, \cite{li2019large} studied transferable features with the hierarchy which encodes the semantic relations.
Their approach turned out to be applicable to the problem of zero-shot learning as well.
\cite{shen2021model} proposed model agnostic regularisation technique in order to leverage the relationship between graph labels to preserve category neighbourhood.
\section{Method}
\label{sec:method} 
This section explains the details of KGTN-ens.
The method extends the KGTN architecture proposed by \cite{chen2020knowledge}, which relies on graph-based knowledge transfer to yield state-of-the-art results on few-show image classification.
The most important difference relies on the usage of multiple graphs instead of a single one, which enables the usage of different knowledge sources.
Each of these graphs generates different prototypes, which are later combined and compared against the output of the feature extractor.
It might be not immediately obvious why the approach with multiple knowledge graphs is used, as they may be merged into one using \texttt{owl:sameAs} or similar property.
Notice that this method does not require knowledge graphs in a strict sense -- KGTM processes only distances between classes, which are later used for scoring prototypes.
Therefore, integrating different sources of knowledge is fairly easy and requires a minimum amount of effort -- the KGTN-ens architecture seamlessly handles different types of distances derived from embeddings. 

\newcommand{\norm}[1]{\left\lVert#1\right\rVert}
\textbf{Problem formulation.}
Following \cite{chen2020knowledge}, the classification task is formulated as learning the prototypes of considered classes.
In the typical approach to classification, the model prediction $\hat{y}$ based on the input $x$ is obtained in the following way:
\begin{equation}
    \hat{y} = \underset{k}{\arg\max} p(y = k | x)
\end{equation}
where $p$ is calculated using the standard softmax function:
\begin{equation}
    p(y = k | x) = \frac{\exp\left( f_{k}(\mathbf{x}) \right)}{\sum_{i=1}^{K} \exp\left( f_{i}(\mathbf{x}) \right)},
\end{equation}
where $K$ is the number of considered classes and $f_{k}$ is the linear classifier.
Since 
\begin{equation}
    \underset{k}{\arg\max} p(y = k | \mathbf{x}) = \underset{k}{\arg\max} f_{k}(\mathbf{x}),    
\end{equation}
the $f_{k}(\mathbf{x})$ can be formulated as follows:
\begin{align}
    f_{k}(\mathbf{x}) &= \mathbf{w}^T_k \mathbf{x} + b_k \\
    &= - \frac{1}{2} \norm{\mathbf{w}_k - \mathbf{x} }^2_2 + \frac{1}{2} \norm{\mathbf{w}_k}^2_2 + \frac{1}{2} \norm{\mathbf{x} }^2_2 + b_k \nonumber
\end{align}
setting $b_k = 0$ and $ \norm{ \mathbf{w}_i }_2 = \lVert \mathbf{w}_j \rVert_2$ for each $i,j$, the classifier $f_k{\mathbf{x}}$ can be perceived as a similarity measure between the extracted features and prototypes:
\begin{align}
    \hat{y} = \underset{k}{\arg\max} p(y = k | \mathbf{x}) = \underset{k}{\arg\min} \norm{\mathbf{w}_k - \mathbf{x} }^2_2.
\end{align}
As a result, $\mathbf{w}_k$ can be interpreted as a prototype for class $k$, and these prototypes are learned during the training process.

\begin{figure*}
    \input{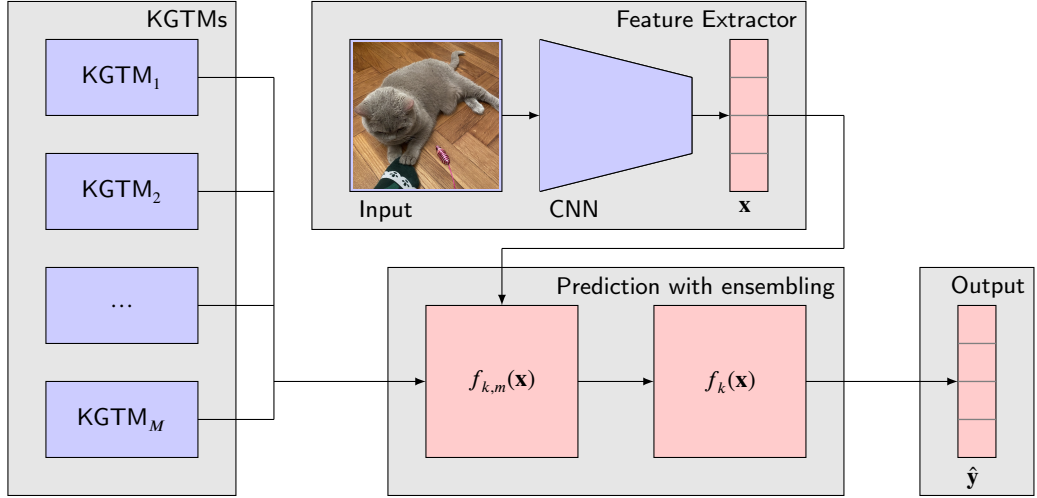}
    \caption{Architecture of KGTN-ens.}
    \label{fig:architecture}
\end{figure*}

The overall architecture of KGTN-ens is presented in Figure \ref{fig:architecture} and it consists of three main parts: \emph{Feature Extractor}, \emph{KGTMs}, and \emph{Prediction with ensembling}.
Feature Extractor is a convolutional neural network that extracts features from the input image, such as ResNet \citep{he2016deep}.
KGTMs refer to the list of knowledge graph transfer modules (each one handles a different knowledge graph) that are used to generate prototypes.
Finally, prediction with ensembling a module that scores extracted features against obtained prototypes in order to make the final classification.

\textbf{KGTMs.}
Since we use the plain ResNet50 for the feature extractor part, we start the description with the KGTMs part.
Consider a dataset of images, where each of them is associated with either a base class or a novel class.
There are $K_{\texttt{base}}$ base classes and $K_{\texttt{novel}}$ novel classes ($K=K_{\texttt{base}}+K_{\texttt{novel}}$).
In the original KGTN approach, the correlations between categories are encoded in a graph $\mathcal{G}=\{ \mathbf{V}, \mathbf{A} \}$, where $\mathbf{V} = \{ v_1, v_2, \dots, v_{K_{\texttt{base}}} \dots, v_K \}$ represents classes and $\mathbf{A}$ denotes an adjacency matrix, in which $A_{i,j}$ is the \emph{correlation} between classes $v_i$ and $v_j$.
Our approach extends this concept in a way in which there are multiple graphs $\mathcal{G}_1, \dots, \mathcal{G}_M$.
Specifically, each of them shares the same classes $\mathbf{V}$ but has different correlation values stored in $\mathbf{A}$ matrices.

Just as KGTN, KGTN-ens is based on Gated Graph Neural Network \citep{li2016gated}, in which each class is represented by a node $v_k$ is associated with a hidden state $h^t_k$ at time $t$.
It is initialised with $\mathbf{h}_k^0=\mathbf{w}_{k}^{\text{init}}$, where $\mathbf{w}_{k}^{\text{init}}$ are chosen at random.
The parameter vector $\mathbf{a}_k^t$ for node $k$ at time $t \in \{1, \ldots, T\}$ is defined as:
\begin{equation}
    \mathbf{a}_k^t=\left[\sum_{k'=1}^K{a_{kk'}\mathbf{h}_{k'}^{t-1}}, \sum_{k'=1}^K{a_{k'k}\mathbf{h}_{k'}^{t-1}}\right],
 \end{equation}
where $a_{kk'}$ denotes the correlation between nodes $k$ and $k'$.
The hidden states $\mathbf{h}_{k}^{t}$ for weight $k$ at time $t$ are determined with a gating mechanism inspired by GRU (abbr. from gated recurrent unit), which was introduced by \cite{cho2014properties}:
\begin{equation}
    \begin{split}
     \mathbf{z}_k^t=&{}\sigma(\mathbf{W}^z{\mathbf{a}_k^t}+\mathbf{U}^z{\mathbf{h}_k^{t-1}}), \\
     \mathbf{r}_k^t=&{}\sigma(\mathbf{W}^r{\mathbf{a}_k^t}+\mathbf{U}^r{\mathbf{h}_k^{t-1}}), \\
     \widetilde{\mathbf{h}_k^t}=&{}\tanh\left(\mathbf{W}{\mathbf{a}_k^t}+\mathbf{U}({\mathbf{r}_k^t}\odot{\mathbf{h}_k^{t-1}})\right), \\
     \mathbf{h}_k^t=&{}(1-{\mathbf{z}_k^t}) \odot{\mathbf{h}_k^{t-1}}+{\mathbf{z}_k^t}\odot{\widetilde{\mathbf{h}_k^t}}.
    \end{split}
 \end{equation}
Here, $\mathbf{W}^z$ and $\mathbf{U}^z$ are the weights for the update gate, and $\mathbf{W}^r$ and $\mathbf{U}^r$ are the weights for the reset gate.
The hyperbolic tangent function is given by tanh, whereas $\sigma$ is the sigmoid function.
The final weight $\mathbf{w}_k^{*}$ for class $k$ is defined as:
\begin{equation}
    \mathbf{w}_k^{*} = o ( \mathbf{h}_k^{T}, \mathbf{h}_k^{0} ),
\end{equation}
where $o$ is the fully connected layer.

\textbf{Prediction and ensembling.}
The classifier $f({\mathbf{x}})$ is treated as a similarity metric between the output of the feature extractor and the most similar class prototypes learned by the knowledge graph transfer module.
In the original KGTN approach, the relationship between these two was calculated using the inner product, cosine similarity or Person's correlation coefficient.
For the inner product, which was the most effective, the classifier was defined as $f_{k}(\mathbf{x}) = \mathbf{x} \cdot \mathbf{w}_{k}^{*}$, where $x$ is the feature vector of an image and $\mathbf{w}_{k}^{*}$ denotes the learned weight for class $k$.
Conventionally, $f({\mathbf{x}}) = \underset{k}{\arg \max} f_{k}(\mathbf{x})$.
However, in our approach, we use the ensembling-inspired technique to improve the performance of the classifier.

In KGTN-ens, we calculate similarity for each of the $m$ available graphs.
Using a similar inner product approach, this is done the following way: $f_{k,m}(\mathbf{x}) = \mathbf{x} \cdot \mathbf{w}_{k,m}^{*}$, where $\mathbf{w}_{k,m}^{*}$ is the learned weight for the $m$-th graph.
Then, the final result for class $k$ has to be chosen.
Such an approach is inspired by ensemble learning strategies, though we do not use \emph{weak learners} in a strict sense.
One of the main drawbacks of ensemble learning -- the linear memory complexity with the proportional computational burden -- is partially avoided, as only the part of the network is multiplied.
Most importantly, the feature extractor, which often can be the largest component of modern architectures, is used only once.
This enables us to fit several knowledge sources on proprietary GPUs (we used a single NVIDIA RTX 2080 Ti in our experiments).
We propose two simple approaches for selecting the final result: mean and maximum.
For the former, the result for class $k$ is the mean of the $m$ products:
\begin{equation}
    f_{k}(\mathbf{x}) = \frac{1}{M} \sum_{m=1}^{M} f_{k,m}(\mathbf{x}).
\end{equation}
In ensemble learning literature, this would be called \emph{soft voting}.
The maximum approach is very similar:
\begin{equation}
    f_{k}(\mathbf{x}) = \max_{m=1}^{M} \left( f_{k,m}(\mathbf{x}) \right),
\end{equation}
In other words, we take the maximum of the similarities for each of the $m$ available graphs.

\textbf{Optimisation.}
To enable fair comparison, we use a two-step training regime similar to \cite{hariharan2017low} and \cite{chen2020knowledge} -- the first is devoted to the feature extractor, whereas the second one fine-tunes the graph-related part of the network.
In the first stage, we train the feature extractor $\phi(\cdot)$ using the base classes from $\mathcal{D}_{base}$.
The loss $\mathcal{L}_1$ calculated in this step consists of the standard cross-entropy loss and squared gradient magnitude loss \citep{hariharan2017low}, which acts as a regularisation term:
\begin{equation}
    \mathcal{L}_1 = \mathcal{L}_c + \lambda \mathcal{L}_s,
\end{equation}
where:
\begin{align}
    \mathcal{L}_c &= - \frac{1}{N_\text{base}} \sum_{i=1}^{N_\text{base}} \sum_{i=1}^{K_\text{base}} \mathbb{1}_{k = y_i} \log p_i^k,\\
    \mathcal{L}_s &= \frac{1}{N_\text{base}} \sum_{i=1}^{N_\text{base}} \sum_{i=1}^{K_\text{base}} \left( p_i^k - \mathbb{1}_{k = y_i} \right)  \norm{\mathbf{x}_i }^2_2,
\end{align}
where $\mathbb{1}$ is the indicator function and $\lambda$ is a loss balance parameter.
In the second stage, the weights of the feature extractor are frozen.
Other parts of the architecture are trained using base and novel samples with the following loss:
\begin{equation}
    \mathcal{L}_2 = - \frac{1}{N} \sum_{i=1}^{N} \sum_{i=1}^{K} \mathbb{1}_{k = y_i} \log p_i^k + \eta \sum_{k=1}^{K} \norm{\mathbf{w}_k^{*} }^2_2,
\end{equation}
where $\eta$ balances the loss components.
\section{Evaluation}
\label{sec:results}

This section contains the results of the conducted experiments.
First, we introduce the used knowledge sources -- semantic similarity graph, WordNet and Wikidata.
Then, we describe the evaluation of KGTN-ens with different combinations of embeddings and compare them with the previous work.
Finally, we provide a detailed analysis and ablation studies.

\subsection{Knowledge sources}
\label{sec:results_knowledge}

In our evaluation, we use three different sources of knowledge, which can be the backbone of KGTMs: \emph{hierarchy}, \emph{glove}, and \emph{wiki}.
The first two have been proposed by \cite{chen2020knowledge}.
The wiki graph is constructed on top of Wikidata, a collaborative knowledge graph connected to Wikipedia \citep{vrandevcic2014wikidata}.
In this subsection, we discuss the preparation of these knowledge sources in detail.

\textbf{Semantic similarity graph (\emph{glove}).}
The first source of knowledge is built from GLoVe word embeddings \citep{pennington2014glove}.
For two words $w_i$ and $w_j$, their semantic distance $d_{i,j}$ is defined as the Euclidean distance between their GLoVe embeddings $\mathbf{f}_i^w$ and $\mathbf{f}_j^w$.
Following \cite{chen2020knowledge}, the final correlation coefficient $a_{i,j}$ is obtained using the following function:
\begin{equation}
    \lambda^{d_{i,j} - \min d_{i,k}, \forall k \neq i},
    \label{eq:monotonic_function}
\end{equation}
where $\lambda=0.4$ and $a_{ii}=1$.

\textbf{WordNet category distance (\emph{hierarchy}).}
This source of knowledge is built from the WordNet hierarchy -- a popular lexical database of English \citep{miller1995wordnet}.
Since ImageNet classes are based on WordNet, the WordNet hierarchy can be used to measure the distance between two classes.
This time the distance $d_{i,j}$ is defined as the number of common ancestors of the two words (categories) $w_i$ and $w_j$.
The output is processed similarly to Equation \eqref{eq:monotonic_function}, except that the $\lambda$ parameter is set to 0.5.

\textbf{Wikidata embeddings (\emph{wiki}).}
The last source of knowledge is built from the Wikidata embeddings.
The mapping between the ImageNet classes and Wikidata is provided by \cite{filipiak2021mapping}.
Having the mapping, the class-corresponding entities from Wikidata can be embedded and used as a class prototypes.
Although there exist some datasets of Wikidata embeddings, they are often incomplete.
Most importantly, they does not contain all the embeddings of ImageNet classes.
Wembedder \citep{nielsen2017wembedder} offers 100-dimensional Wikidata embeddings made using the word2vec algorithm \citep{mikolov2013efficient}, but it bases on an incomplete dump of Wikidata and does not contain all the classes nedded in the ImageNet-FS dataset.
\cite{zhu2019graphvite} proposed Graphvite, a general graph embedding engine.
\textit{Wikidata5m} is a large dataset of 5 million Wikidata entities, which is used to train the embeddings.
The framework comes with embeddings created using numerous popular algorithms, such as TransE, DistMult, ComplEx, SimplE, RotatE, and QuatE.
However, 891 out of 1000 entities used in the ImageNet are embedded, which was not enough for performing the experiment.

We used the pre-trained 200-dimensional embeddings of Wikidata entities from PyTorch BigGraph \citep{lerer2019pytorch}, which are publicly available\footnote{\url{https://torchbiggraph.readthedocs.io/en/latest/pretrained_embeddings.html}}.
The embeddings were prepared using the full Wikidata dump from 2019-03-06.
All but three entities were directly mapped to embeddings to their Wikidata ID.
Three entities (\texttt{Q1295201}, \texttt{Q98957255}, \texttt{Q89579852}) could not be instantly matched -- they were manually matched to \texttt{"grocery store"@en}, \texttt{"cricket"@en}, and \texttt{Q655301} respectively.
Having the mapping, now we create an embedding array, ordered as the mappings in the original KGTN paper (that is, as a $1000 \times 200$ array, where 200 denotes the dimensionality of a single embedding).
The same function from Equation~\eqref{eq:monotonic_function} was used to generate final correlations between the embeddings, although this time $\lambda=0.32$ was used (see Section \ref{sec:details_ablation}).

\subsection{Experiment results}
\label{sec:results_experiment}

In this subsection, we present the results of the conducted experiments.
We describe the evaluation data -- the experiment has been conducted on ImageNet-FS dataset.
The training hyperparameters and the setup is also described.
We also describe the evaluation protocol, as well as the evaluation metrics.
Finally, we present the results of the experiments and compare them with the previous work.

\textbf{Data.}
Similarly to Chen et al., our approach has been evaluated on ImageNet-FS, a popular benchmark for few-shot learning task.
ImageNet-FS contains 1,000 classes from ImageNet Large Scale Visual Recognition Challenge 2012 \citep{ILSVRC15}, of which 389 belongs to the base category and 611 to the novel category.
193 base classes and 300 novel classes are used for training and cross-validation, whereas the test phase is performed on the remaining 196 categories and 311 novel classes.
Base categories consist of around 1280 train and 50 test images per each class.
The authors of KGTN also evaluated their solution against a larger dataset, ImageNet-6K, which contains 6,000 classes (of which 1,00 belongs to the novel category).
Unfortunately, we were unable to test KGTN-ens using this dataset, since it has not been made public nor available to us at the time of writing this paper.

\textbf{Training.}
To enable fair comparison, we used the same 2-step training and evaluation procedures as in KGTN.
Stochastic gradient descent (SGD) was used to train the model with a batch size equal to 256 (divided equally for base and novel classes), a momentum of 0.9, and a weight decay of 0.0005.
The learning rate is initially set at 0.1 and divided by 30 at every 30 epochs.
In general, we used the same hyperparameters as in KGTN unless stated otherwise.

\textbf{Setup.}
All the experiments have been conducted on a single NVIDIA GeForce RTX 2080 Ti GPU.
We used the code released by the authors of KGTN and modified it to support the KGTN-ens approach.
PyTorch \cite{paszke2017automatic} was used to conduct the experiments.
The code will be released after the publication of this article.

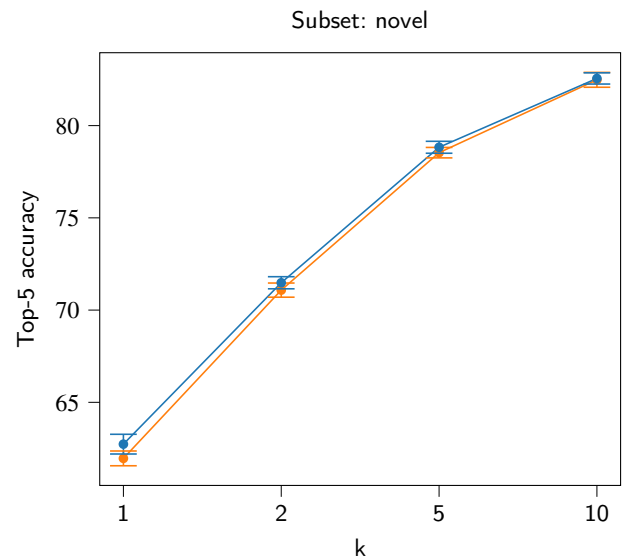
\begin{figure}
    \centering    
    \begin{subfigure}[b]{.5\textwidth}
        \centering
\begin{tikzpicture}

\definecolor{darkgray176}{RGB}{176,176,176}
\definecolor{darkorange25512714}{RGB}{255,127,14}
\definecolor{lightgray204}{RGB}{204,204,204}
\definecolor{steelblue31119180}{RGB}{31,119,180}

\begin{axis}[
legend cell align={left},
legend style={
  fill opacity=0.8,
  draw opacity=1,
  text opacity=1,
  at={(0.97,0.03)},
  anchor=south east,
  draw=lightgray204
},
tick align=outside,
tick pos=left,
title={Subset: novel},
x grid style={darkgray176},
xlabel={k},
xmin=-0.15, xmax=3.15,
xtick style={color=black},
xtick={0,1,2,3},
xtick={0,1,2,3},
xtick={0,1,2,3},
xtick={0,1,2,3},
xtick={0,1,2,3},
xtick={0,1,2,3},
xticklabels={1,2,5,10},
xticklabels={1,2,5,10},
xticklabels={1,2,5,10},
xticklabels={1,2,5,10},
xticklabels={1,2,5,10},
xticklabels={1,2,5,10},
y grid style={darkgray176},
ylabel={Top-5 accuracy},
ymin=60.4949736662386, ymax=83.9517212149162,
ytick style={color=black}
]

\path [draw=darkorange25512714, semithick]
(axis cs:0,61.5611894639058)
--(axis cs:0,62.3616401180878);

\path [draw=darkorange25512714, semithick]
(axis cs:1,70.6993290183528)
--(axis cs:1,71.4665873803611);

\path [draw=darkorange25512714, semithick]
(axis cs:2,78.2504533010468)
--(axis cs:2,78.8170708147088);

\path [draw=darkorange25512714, semithick]
(axis cs:3,82.0791248078314)
--(axis cs:3,82.885505417249);

\path [draw=steelblue31119180, semithick]
(axis cs:0,62.1986523071615)
--(axis cs:0,63.2675856349607);

\path [draw=steelblue31119180, semithick]
(axis cs:1,71.1533675453303)
--(axis cs:1,71.8099765061167);

\path [draw=steelblue31119180, semithick]
(axis cs:2,78.502138954217)
--(axis cs:2,79.1493079911206);

\path [draw=steelblue31119180, semithick]
(axis cs:3,82.2532127531485)
--(axis cs:3,82.8606136134109);

\addplot [semithick, darkorange25512714, mark=-, mark size=5, mark options={solid}, only marks]
table {%
0 61.5611894639058
1 70.6993290183528
2 78.2504533010468
3 82.0791248078314
};
\addplot [semithick, darkorange25512714, mark=-, mark size=5, mark options={solid}, only marks]
table {%
0 62.3616401180878
1 71.4665873803611
2 78.8170708147088
3 82.885505417249
};
\addplot [semithick, darkorange25512714, mark=*, mark size=1.5, mark options={solid}]
table {%
0 61.9614147909968
1 71.0829581993569
2 78.5337620578778
3 82.4823151125402
};
\addplot [semithick, steelblue31119180, mark=-, mark size=5, mark options={solid}, only marks]
table {%
0 62.1986523071615
1 71.1533675453303
2 78.502138954217
3 82.2532127531485
};
\addplot [semithick, steelblue31119180, mark=-, mark size=5, mark options={solid}, only marks]
table {%
0 63.2675856349607
1 71.8099765061167
2 79.1493079911206
3 82.8606136134109
};
\addplot [semithick, steelblue31119180, mark=*, mark size=1.5, mark options={solid}]
table {%
0 62.7331189710611
1 71.4816720257235
2 78.8257234726688
3 82.5569131832797
};
\end{axis}

\end{tikzpicture}
        \label{fig:max_novel}
    \end{subfigure}  
    \caption{KNGT-ens (blue) performance mean top-5 accuracy with compared to the KGTN (orange) over 5 runs. KGNT-ens uses glove and hierarchy graphs combined with the max ensembling function. Horizontal lines indicate standard deviations. }
    \label{fig:kgtn-ens_performance}
\end{figure}

\textbf{Evaluation.}
Following previous work in few-shot learning, we report our evaluation results in terms of the top-5 accuracy of novel and all (base + novel) classes in the $k$-shot learning task, where $k \in \{1,2,5, 10\}$ is the number of classes in the novel category.
Following \cite{hariharan2017low} and \cite{chen2020knowledge}, we repeat each experiment five times and report the averaged values of the top-5 accuracy.
Table~\ref{tab:results} shows the classification results compared with some of the recent state-of-the-art benchmarks.
Figure \ref{fig:kgtn-ens_performance} presents the top-5 accuracy of the KGTN-ens model on ImageNet-FS.
Of three possible combinations of the three sources of knowledge, the KGTN-ens model performed best with the combination of \emph{hierarchy} and \emph{glove}.
Notably, it performed better than KGTN with these two sources of knowledge alone.
Compared to KGTN (with inner product similarity and glove embeddings), the KGTN-ens model (inner product, max ens. function, glove and hierarchy embeddings) achieved +0.63, +0.58, +0.43, +0.26 pp. top-5 accuracy on novel classes for $k \in \{ 1, 2, 5, 10\}$ respectively.
The smaller the $k$, the higher the performance gain.
It also beats the more recent graph-based framework proposed by \cite{shen2021model} by +1.73/+1.18/+0.20 pp. top-5 accuracy on novel classes.
For the all classes, the KGTN-ens model achieved +0.26, +0.25, +0.32, --0.04 pp. top-5 accuracy compared to the same KGTN model for $k \in \{ 1, 2, 5, 10\}$ respectively.

\begin{table*}
    \centering
    \caption{Top-5 accuracy on \emph{novel} and \emph{all} subsets on ImageNet-FS. All the methods used ResNet-50 as a feature extractor. Partially based on the data provided by \cite{chen2020knowledge}.}
    \label{tab:results}
    \begin{tabular}{l rrrr rrrr}
    \toprule
     & \multicolumn{4}{c}{novel} & \multicolumn{4}{c}{all} \\ \cmidrule(lr){2-5} \cmidrule(lr){6-9}
     & \multicolumn{1}{c}{1} & \multicolumn{1}{c}{2} & \multicolumn{1}{c}{5} & \multicolumn{1}{c}{10} & \multicolumn{1}{c}{1} & \multicolumn{1}{c}{2} & \multicolumn{1}{c}{5} & \multicolumn{1}{c}{10} \\
    \midrule
    MN \cite{vinyals2016matching} & 53.5 & 63.5 & 72.7 & 77.4 & 64.9 & 71.0 & 77.0 & 80.2 \\
    PN \cite{snell2017prototypical} & 49.6 & 64.0 & 74.4 & 78.1 & 61.4 & 71.4 & 78.0 & 80.0 \\
    SGM \cite{hariharan2017low} & 54.3 & 67.0 & 77.4 & 81.9 & 60.7 & 71.6 & 80.2 &  83.6 \\
    SGM w/ G \cite{hariharan2017low} & 52.9 & 64.9 & 77.3 & 82.0 & 63.9 & 71.9 & 80.2 & 83.6 \\
    AWG \cite{gidaris2018dynamic} & 53.9 & 65.5 & 75.9 & 80.3 & 65.1 & 72.3 & 79.1 & 82.1 \\
    PMN \cite{wang2018low} & 53.3 & 65.2 & 75.9 & 80.1 & 64.8 & 72.1 & 78.8 & 81.7 \\
    PMN w/ G \cite{wang2018low} & 54.7 & 66.8 & 77.4 & 81.4 & 65.7 & 73.5 & 80.2 & 82.8 \\
    LSD \cite{douze2018low} & 57.7 & 66.9 & 73.8 & 77.6 & -- & -- & -- & -- \\
    KTCH \cite{li2019large} & 58.1 & 67.3 & 77.6 & 81.8 & -- & -- & -- & -- \\
    IDeMe-Net \cite{chen2019image} & 60.1 & 69.6 & 77.4 & 80.2 & -- & -- & -- & -- \\
    KGTN-CosSim \cite{chen2020knowledge} & 61.4 & 70.4 & 78.4 & 82.2 &  67.7 & 74.7 & 80.9 & \textbf{83.6} \\
    KGTN-PearsonCorr \cite{chen2020knowledge} & 61.5 & 70.6 & 78.5 & 82.3 & 67.5 & 74.4 & 80.7 & 83.5 \\
    KGTN-InnerProduct \cite{chen2020knowledge} & 62.1 & 70.9 & 78.4 & 82.3 & 68.3 & 75.2 & 80.8 & 83.5 \\
    SGM with graph regularisation \cite{shen2021model} & 61.1 & 70.3 & 78.6 & -- & -- & -- & -- & -- \\
    KGTN-ens (ours) & \textbf{62.73} & \textbf{71.48} & \textbf{78.83} & \textbf{82.56} & \textbf{68.58} & \textbf{75.45} & \textbf{81.12} & 83.46 \\
    \bottomrule
    \end{tabular}    
\end{table*}

\subsection{Details and ablation studies}
\label{sec:details_ablation}

This subsection provides more details on the KGTN-ens model and its ablation studies.
We analyse the impact of the following factors on the performance of the KGTN-ens model: adjacency matrices, used embeddings, ensembling method, similarity function, and variance of the results.

\begin{figure}%
    \begin{subfigure}{.5\textwidth}
        \centering
        \label{fig:ablation_dist_raw}
        \input{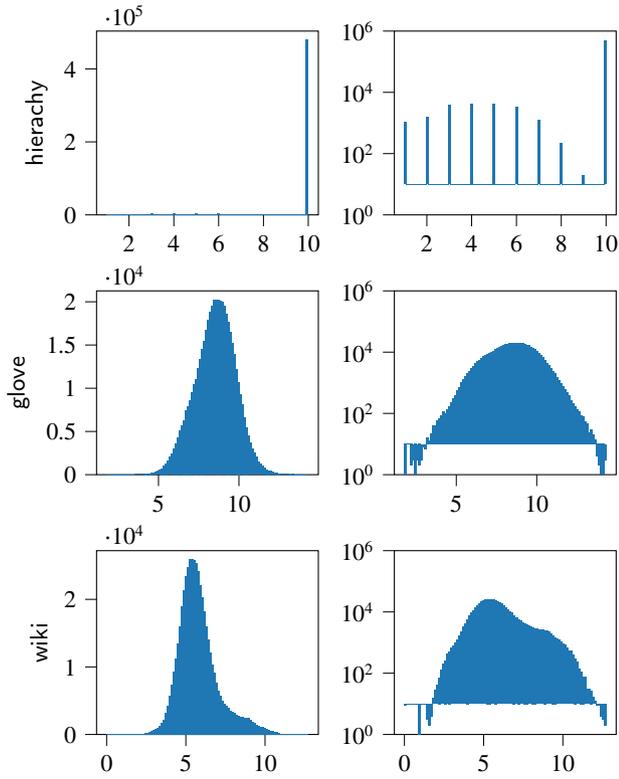}
        \caption{Raw values}
    \end{subfigure}
    \begin{subfigure}{.5\textwidth}
        \centering
        \label{fig:ablation_dist_processed}
        \input{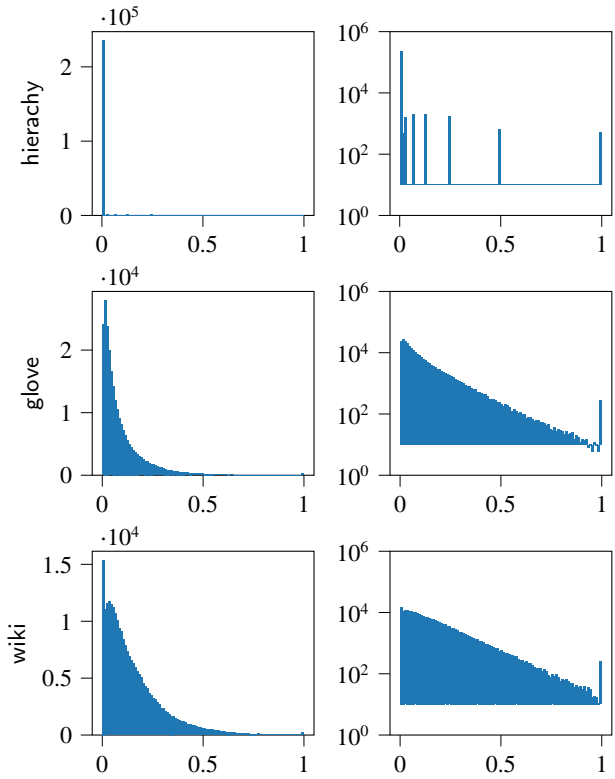}
        \caption{Processed values}
    \end{subfigure}  
    \caption{Adjacency matrix distributions}
    \label{fig:ablation_dist}
\end{figure}

\textbf{Adjacency matrix analysis.}
Since glove knowledge graph was the most effective for KGTN, we assume that wiki should roughly resemble it in terms of its distribution.
In order to investigate the similarity between distributions, adjacency matrices have been created using pairwise euclidean distances.
While glove and wiki are normal-like, the distribution for hierarchy is bimodal and most of the distances are the highest ones (Figure \ref{fig:ablation_dist}).
To assess the correlation between adjacency matrices, Mantel tests have been performed (Table \ref{tab:mantel_raw}).
The values marked as processed were run through Equation \eqref{eq:monotonic_function}.
Correlations of the processed matrices are visibly higher compared to raw ones, especially regarding glove and wiki).
The highest correlation has been observed between glove and wiki.

\begin{table}
    \centering
    \caption{Descriptive statistics about the adjacency matrices.}
    \label{tab:ablation_adjacency_stats}
    \begin{tabular}{ll rrrr}
    \toprule
    & KG & Min & Avg & Max & Std \\
    \midrule
    \multirow{3}{*}{Raw} & hierarchy & 0.00 & 9.76 & 10.00 & 1.20 \\
    & glove & 0.00 & 8.52 & 14.31 & 1.29 \\
    & wiki & 0.00 & 5.82 & 12.73 & 1.32 \\ \cmidrule(){2-6}
    \multirow{3}{*}{Processed} & hierachy   & 0.00 & 0.01 & 2.00 & 0.07 \\
    & glove      & 0.00 & 0.05 & 2.00 & 0.11 \\
    & wiki       & 0.00 & 0.08 & 2.00 & 0.14 \\ \bottomrule
\end{tabular}
\end{table}

\begin{table}
    \caption{Mantel test results.}
    \label{tab:mantel_raw}
    \begin{tabular}{lll rr}
        \toprule
        & KG$_{1}$ & KG$_{1}$ & correlation & p-value \\ \midrule
        \multirow{3}{*}{Raw} & hierachy   & glove      & 0.14 & 0.001 \\
        & hierachy   & wiki       & 0.13 & 0.001 \\
        & glove      & wiki       & 0.16 & 0.001 \\ \cmidrule(){2-5}
        \multirow{3}{*}{Processed} & hierachy   & glove & 0.19 & 0.001 \\
        & hierachy   & wiki       & 0.18 & 0.001 \\
        & glove      & wiki       & 0.44 & 0.001 \\ \bottomrule
    \end{tabular}
\end{table}

\textbf{Importance of used knowledge graphs.}
Firstly, we analyse the influence of the used KGs separately (without ensembling) -- that is, with the original KGTN architecture.
Table \ref{tab:ablation_kgtn_graphs_and_functions} shows the results of the ablation studies on the three knowledge graphs.
The hierarchy and glove knowledge graphs are the ones examined by \cite{chen2020knowledge}, whereas the wiki knowledge graph is the one introduced in our experiments.
In order to ensure that the advantage comes from the knowledge encoded in KGs, Chen et al. argue that glove and hierarchy embeddings perform better than uniform (all correlations set to $1/K$) and random (correlations drawn from the uniform distributions) distance matrices.
Similarly, the usage of wiki knowledge graph yielded generally better results (up to +3.44 pp for 1-shot in the novel category) compared to random and uniform cases, which constitutes a noticeable improvement.
However, compared to glove and hierarchy, the wiki knowledge graph yields worse results -- notably for low-shot scenarios.
We hypothesise that the difference in the performance of wiki knowledge graph is due to the low quality of embeddings, as some issues regarding their accuracy were previously reported\footnote{\url{https://datascience.stackexchange.com/q/95007/8949}}.


\begin{table*}
    \caption{Classification results for KGTN on different embeddings tested against different similarity functions. Results for glove and hierarchy embeddings are provided by \cite{chen2020knowledge}. Bold results mean the best ones for the wiki graph only.}
    \label{tab:ablation_kgtn_graphs_and_functions}
    \begin{tabular}{ll rrrr rrrr}
        \toprule
        & & \multicolumn{4}{c}{novel} & \multicolumn{4}{c}{all} \\ \cmidrule(lr){3-6} \cmidrule(lr){7-10}
        knowledge graph & similarity function & \multicolumn{1}{c}{1} & \multicolumn{1}{c}{2} & \multicolumn{1}{c}{5} & \multicolumn{1}{c}{10} & \multicolumn{1}{c}{1} & \multicolumn{1}{c}{2} & \multicolumn{1}{c}{5} & \multicolumn{1}{c}{10} \\
        \midrule
        \multirow{3}{*}{wiki} & cosine similarity   &  56.65 &  \textbf{68.21} &  77.31 &  81.88 &  64.59 &  73.32 &  80.03 &  \textbf{83.44} \\
        & inner product &  55.55 &  67.81 &  \textbf{77.99} &  \textbf{82.15} &  \textbf{64.61} &  \textbf{73.28} &  \textbf{80.55} &  83.22 \\
        & Pearson correlation &  \textbf{56.84} &  68.10 &  77.03 &  81.62 &  64.03 &  72.61 &  79.53 &  83.20 \\ \cmidrule(lr){2-10}
        \multirow{3}{*}{glove} & cosine similarity & 61.4 & 70.4 & 78.4 & 82.2 &  67.7 & 74.7 & 80.9 & 83.6 \\
        & inner product & 62.1 & 70.9 & 78.4 & 82.3 & 68.3 & 75.2 & 80.8 & 83.5 \\
        & Pearson correlation & 61.5 & 70.6 & 78.5 & 82.3 & 67.5 & 74.4 & 80.7 & 83.5 \\ \cmidrule(lr){2-10}
        \multirow{1}{*}{hierarchy} & inner product & 60.1 & 69.4 & 78.1 & 82.1 & 67.0 & 74.4 & 80.7 & 83.3\\ \cmidrule(lr){2-10}
        \multirow{1}{*}{(uniform)} & inner product & 53.4 & 67.4 & 78.8 & 81.5 & 63.8 & 73.3 & 80.3 & 82.9 \\ \cmidrule(lr){2-10}
        \multirow{1}{*}{(random)} & inner product & 54.4 & 67.4 & 77.8 & 81.9 & 64.5 & 73.3 & 80.5 & 83.2 \\ 
    \bottomrule
    \end{tabular}
\end{table*}

\textbf{Importance of the ensembling method.}
Table \ref{tab:results_ensembling} presents results for the different ensembling strategies compared to the KGTN baseline, which can be treated as a KGTN-ens model with no ensembling.
Mean ensembling gave mixed results compared to the baseline ($+0.34$, $-0.63$, $+0.36$, $-0.27$ pp. for novel classes and $-1.45$, $-1.41$, $+0.14$, $-0.17$ pp. for all classes, both groups for $k \in \{1, 2, 5, 10\}$ respectively.
However, using the max ensembling strategy has been better in all the cases ($+0.77$, $+0.40$, $+0.29$, $+0.07$ pp. for novel classes and $+0.24$, $+0.18$, $+0.19$, $+0.06$ pp. for all classes).
A possible explanation of this effect might stem from the \emph{winner takes all} nature of the maximum function, which chooses the most similar embedding to the given prototype and rejects other, potentially improper, embeddings.
At the same time, these improper embeddings still contribute to the overall formula for the mean ensembling function.
However, research on a larger number of employed knowledge graphs has to be conducted to validate this hypothesis.

\textbf{Variance of the results.}
Contrary to expectations, adding additional knowledge sources slightly increase the variance of the results in most cases (Table \ref{tab:stddev}).
A possible explanation of these results is the fact that KGTN-ens is not an ensembling technique in the typical sense of this word, but rather a way of choosing the embeddings of the different knowledge sources.
We report results for novel classes only, as the difference in variance is amplified among these (see also Fig. \ref{fig:kgtn-ens_performance}).
No significant differences in the variance of mean and max ensembling have been found.
The variance of the results for baseline KGTN has been obtained using five runs of the original KGTN with glove embeddings.

\begin{table}
    \centering
    \caption{Standard deviations of top-5 accuracy. Abbreviations: h -- hierarchy, g -- glove, w -- wiki.}
    \label{tab:stddev}
    \begin{tabular}{llrrrr}
    \toprule
     & & \multicolumn{4}{c}{novel} \\
    type & KG & \multicolumn{1}{c}{1} & \multicolumn{1}{c}{2} & \multicolumn{1}{c}{5} & \multicolumn{1}{c}{10} \\
    \midrule
    KGTN (baseline) & g & 0.40 & 0.38 & 0.28 & 0.40 \\ \cmidrule(lr){2-6}
    \multirow{4}{*}{KGTN-ens (max)} & h+g & 0.53 & 0.33 & 0.32 & 0.30 \\
          & w+g & 0.59 & 0.16 & 0.30 & 0.34 \\
          & w+h & 0.66 & 0.25 & 0.26 & 0.28 \\
          & w+h+g & 0.45 & 0.27 & 0.34 & 0.30 \\  \cmidrule(lr){2-6}
    \multirow{4}{*}{KGTN-ens (mean)} & h+g & 0.57 & 0.31 & 0.31 & 0.34 \\
        & w+g & 0.24 & 0.28 & 0.33 & 0.38 \\
        & w+h & 0.56 & 0.31 & 0.41 & 0.31 \\
        & w+h+g & 0.53 & 0.37 & 0.33 & 0.42 \\
    \bottomrule
    \end{tabular}
    \end{table}

\textbf{Importance of similarity function.}
Table \ref{tab:ablation_kgtn_graphs_and_functions} includes data for performing ablative studies for KGTN with the three different similarity functions: cosine similarity, inner product and Pearson correlation.
\cite[]{chen2020knowledge} analysed all these for KGTN with glove embeddings.
In general, the inner product showed the best performance.
These conclusions can be extrapolated to the wiki graph, as the inner product usually turned out to be the most effective in terms of the top-5 accuracy.
Interestingly, Pearson correlation displayed the best performance for the 1-shot scenario with novel classes.
Table \ref{tab:results_cosine} presents results for the different similarity functions used in the KGTN-ens.
While the combination of hierarchy and glove embeddings was usually the best for cosine similarity as well, the results are visibly worse compared to the inner product similarity function (e.g. $-3.16$ pp. top-5 accuracy difference for 1-shot scenario among novel classes).
Noticeably, the combination of these two graphs and cosine similarity function performed worse than KGTN solely based on glove embeddings (for example, there is a $-2.39$ pp. difference for top-5 accuracy difference for 1-shot scenario among novel classes).

\begin{table*}
    \centering
    \caption{Knowledge graph ensembling (sum), top-5 accuracy, inner product}
    \label{tab:results_ensembling}
    \begin{tabular}{llrrrr rrrr}
    \toprule
    \multicolumn{2}{l}{} & \multicolumn{4}{c}{novel} & \multicolumn{4}{c}{all} \\ \cmidrule(lr){3-6} \cmidrule(lr){7-10}
    \multicolumn{1}{c}{type} & \multicolumn{1}{c}{knowledge graphs} & \multicolumn{1}{c}{1} & \multicolumn{1}{c}{2} & \multicolumn{1}{c}{5} & \multicolumn{1}{c}{10} & \multicolumn{1}{c}{1} & \multicolumn{1}{c}{2} & \multicolumn{1}{c}{5} & \multicolumn{1}{c}{10} \\
    \midrule
    KGTN (baseline) & glove & 61.96 & 71.08 & 78.53 & 82.48 & 68.34 & 75.27 & 80.92 & 83.40   \\ \cmidrule(lr){2-10}
    \multirow{4}{*}{KGTN-ens (mean)} & hierarchy + glove & 62.30 & 70.45 & \textbf{78.90} & 82.21 & 66.89 & 73.86 & 81.06 & 83.22   \\
       & wiki + glove & 60.41 & 69.41 & 78.81 & 82.10 & 66.07 & 73.30 & 81.01 & 83.15   \\
       & wiki + hierarchy + glove & 58.74 & 67.95 & 78.74 & 81.70 & 63.90 & 71.43 & 80.91 & 82.88   \\
       & wiki + hierarchy & 57.89 & 67.49 & 78.49 & 81.91 & 64.10 & 71.83 & 80.80 & 83.04   \\ \cmidrule(lr){2-10}
       \multirow{4}{*}{KGTN-ens (max)} & hierarchy + glove & \textbf{62.73} & \textbf{71.48} & 78.83 & \textbf{82.56} & \textbf{68.58} & \textbf{75.45} & \textbf{81.12} & \textbf{83.46}  \\
       & wiki + glove & 61.21 & 70.66 & 78.60 & 82.34 & 67.69 & 75.04 & 80.95 & 83.33  \\
       & wiki + hierarchy + glove & 61.32 & 70.77 & 78.70 & 82.38 & 67.85 & 75.06 & 81.06 & 83.35  \\
       & wiki + hierarchy & 58.77 & 69.17 & 78.44 & 82.25 & 66.17 & 74.01 & 80.86 & 83.26  \\
    \bottomrule
    \end{tabular}
\end{table*}


\begin{table*}
    \centering
    \caption{Ablation on different similarity functions used in KGTN-ens, top-5 accuracy.}
    \label{tab:results_cosine}
    \begin{tabular}{llrrrrrrrr}
    \toprule
    \multicolumn{2}{l}{} & \multicolumn{4}{c}{novel} & \multicolumn{4}{c}{all} \\ \cmidrule(lr){3-6} \cmidrule(lr){7-10}
    \multicolumn{1}{c}{type} & \multicolumn{1}{c}{knowledge graphs} & \multicolumn{1}{c}{1} & \multicolumn{1}{c}{2} & \multicolumn{1}{c}{5} & \multicolumn{1}{c}{10} & \multicolumn{1}{c}{1} & \multicolumn{1}{c}{2} & \multicolumn{1}{c}{5} & \multicolumn{1}{c}{10} \\
    
    \midrule
       KGTN &                      glove & 61.96 & 71.08 & 78.53 & 82.48 & 68.34 & 75.27 & 80.92 & 83.40 \\\cmidrule(lr){2-10}
       \multirow{4}{*}{KGTN-ens (inner prod.)} & hierarchy + glove & \textbf{62.73} & \textbf{71.48} & \textbf{78.83} & \textbf{82.56} & \textbf{68.58} & \textbf{75.45} & \textbf{81.12} & \textbf{83.46}  \\
       & wiki + glove & 61.21 & 70.66 & 78.60 & 82.34 & 67.69 & 75.04 & 80.95 & 83.33  \\
       & wiki + hierarchy + glove & 61.32 & 70.77 & 78.70 & 82.38 & 67.85 & 75.06 & 81.06 & 83.35  \\
       & wiki + hierarchy & 58.77 & 69.17 & 78.44 & 82.25 & 66.17 & 74.01 & 80.86 & 83.26  \\\cmidrule(lr){2-10}
       \multirow{4}{*}{KGTN-ens (cosine sim.)} &       hierarchy + glove & 59.57 & 69.40 & 77.29 & 81.89 & 64.86 & 73.72 & 80.05 & \textbf{83.46} \\
        &            wiki + glove & 58.34 & 68.75 & 77.24 & 81.84 & 64.43 & 73.44 & 80.01 & 83.38 \\
        &  wiki + hierarchy + glove & 57.75 & 68.44 & 77.20 & 81.90& 63.81 & 73.12 & 79.99 & 83.44 \\
        &        wiki + hierarchy & 57.35 & 68.50 & 77.27 & 81.90  & 63.87 & 73.23 & 80.00 & 83.43 \\

    \bottomrule
    \end{tabular}
    \end{table*}
    
\section{Conclusion}
\label{sec:conclusion}

In this work, we proposed KGTN-ens, which builds on KGTN and allows the incorporation of multiple knowledge sources in order to achieve better performance.
We evaluated KGTN-ens on the ImageNet-FS dataset and showed that it outperforms KGTN in most of the tested settings.
We also evaluated Wikidata embeddings in the same task and showed that they are not as effective as the other embeddings.
We believe that the proposed approach can be used in other few-shot learning tasks and we plan to test it in the future.
Although not publicly available at the time of writing this article, further work might include an evaluation of the proposed approach on ImageNet-6K dataset \cite[]{chen2020knowledge}.
A certain limitation of this study is the fact that it might not scale well for extreme classification problems, due to the calculation of pairwise distances of nodes from large knowledge graphs requiring quadratic memory complexity.
\section*{Acknowledgements}
This research was co-funded by Interreg \"Osterreich-Bayern 2014-2020 programme project KI-Net: Bausteine f\"ur KI-basierte Optimierungen in der industriellen Fertigung (grant agreement: AB 292).

\printcredits

\bibliographystyle{cas-model2-names}

\bibliography{bibliography}

\bio{}
\endbio


\end{document}